%% file: root.tex
\documentclass[journal]{IEEEtran}

\IEEEoverridecommandlockouts                              

\usepackage{graphics} 
\usepackage[pdftex]{graphicx}
\usepackage{epsfig} 
\usepackage{times} 
\usepackage{amsmath} 
\usepackage{amssymb}  
\usepackage{gensymb}
\usepackage{multicol}
\usepackage{multirow}

\usepackage[table]{xcolor}
\usepackage{caption}
\usepackage{subcaption}
\usepackage{array}
\usepackage{hyperref}
\usepackage{cleveref}
\usepackage{xcolor}
\usepackage{soul}
\usepackage{colortbl}

\usepackage{amsmath}
\DeclareMathOperator*{\argmax}{argmax}
\DeclareMathOperator*{\argmin}{argmin}

\begin{document}


\title{Design and Evaluation of a Generic Visual SLAM Framework for Multi Camera Systems}
\author{Pushyami Kaveti$^{1}$, Shankara Narayanan Vaidyanathan $^{2}$, Arvind Thamil Chelvan$^{1}$, Hanumant Singh$^{1}$

\thanks{Manuscript received: June, 23, 2023; Accepted August, 29, 2023.}
\thanks{This paper was recommended for publication by Editor Javier Civera upon evaluation of the Associate Editor and Reviewers' comments.)} 
\thanks{$^{1}$Pushyami Kaveti, Arvind Thamil Chelvan and Hanumant Singh are with College of Engineering, Northeastern University, Boston, Massachusetts, United States of America (email: 
        {\tt\footnotesize \scriptsize \{kaveti.p, thamilchelvan.a, ha.singh\} @northeastern.edu)}}%
\thanks{$^{2} $Shankara Narayanan Vaidyanathan is with Khoury College of Computer Sciences, Northeastern University, Boston, Massachusetts, United States of America (email: {\tt\footnotesize \scriptsize  vaidyanathan.sh@northeastern.edu)}}%
\thanks{Digital Object Identifier (DOI): 10.1109/LRA.2023.3316609}
}

\markboth{IEEE Robotics and Automation Letters. Preprint Version. Accepted September, 2023}
{Kaveti \MakeLowercase{\textit{et al.}}: Design and Evaluation of a Generic Visual SLAM Framework for Multi Camera Systems}

\maketitle

\input{sections/abstract}
\input{sections/introduction}
\input{sections/relatedwork}
\input{sections/slam_framework}
\input{sections/experimental_setup}
\input{sections/results}
\input{sections/conclusion}


\addtolength{\textheight}{-12cm}   





\bibliographystyle{IEEEtran}
\bibliography{references}


\end{document}

%% file: sections/abstract.tex
\begin{abstract}
Multi-camera systems have been shown to improve the accuracy and robustness of SLAM estimates, yet state-of-the-art SLAM systems predominantly support monocular or stereo setups. This paper presents a generic sparse visual SLAM framework capable of running on any number of cameras and in any arrangement. Our SLAM system uses the generalized camera model, which allows us to represent an arbitrary multi-camera system as a single imaging device. Additionally, it takes advantage of the overlapping fields of view (FoV) by extracting cross-matched features across cameras in the rig. This limits the linear rise in the number of features with the number of cameras and keeps the computational load in check while enabling an accurate representation of the scene. We evaluate our method in terms of accuracy, robustness, and run time on indoor and outdoor datasets that include challenging real-world scenarios such as narrow corridors, featureless spaces, and dynamic objects. We show that our system can adapt to different camera configurations and allows real-time execution for typical robotic applications. Finally, we benchmark the impact of the critical design parameters - the number of cameras and the overlap between their FoV that define the camera configuration for SLAM. All our software and datasets are freely available for further research. 
\newline
\end{abstract}

%% file: sections/introduction.tex
\section{Introduction}
\IEEEPARstart{S}{imultaneous} Localization and Mapping (SLAM) is a fundamental task required for the autonomous navigation of mobile robots. 
\begin{figure}[ht!]
\centering
\captionsetup{font={footnotesize}, labelfont={bf,sf}}
\begin{tabular}{c}
\subfloat[]{\includegraphics[ width =0.9\columnwidth]{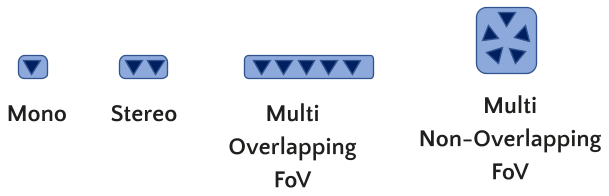}} \\
\subfloat[]{\includegraphics[ width = 0.95\linewidth]{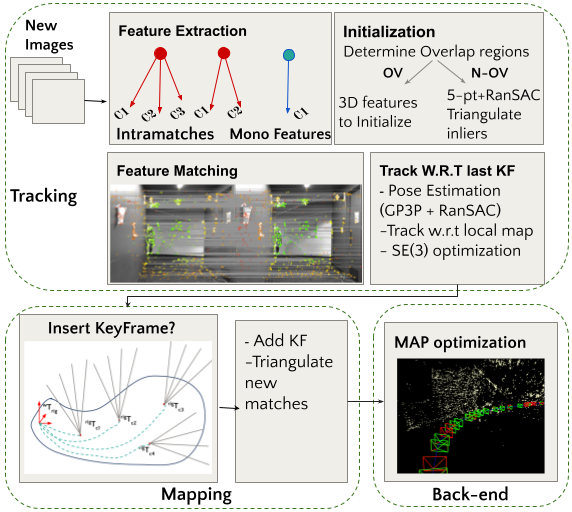}}
\end{tabular}
\caption{(a) Illustration of various overlapping(OV) and non-overlapping(N-OV) camera configurations evaluated in this work. (b) Block diagram of the generic visual SLAM framework showing its sub-modules. Feature extraction computes two types of features\textendash multi-view intra-matches and regular mono features. Note the changes, made to initialization, tracking and keyframe representation to adapt to a general multi-camera system.}
\label{fig:intro_pic}
\end{figure}
Visual SLAM (VSLAM) \cite{davison2003real} is one of the more favored approaches as cameras are inexpensive, easily deployable, low-power sensors that provide rich information about the world in which the robot operates. Most VSLAM pipelines \cite{davison2003real}\cite{campos2021orb}\cite{newcombe2011dtam}\cite{Forster2017} use monocular cameras prone to a single point of failure due to degenerate motion, illumination changes, featureless surfaces, and dynamic objects \cite{he2020review}. They are further limited by a small field of view and inconsistencies associated with scale \cite{engel2015large}\cite{kerl2013robust}. Additional sensors such as Inertial Measurement Units (IMUs) can be used to improve performance in these scenarios, but they can cause large accumulated errors when visual tracking fails. On the other hand, multi-camera systems strike a good balance between data quality and cost. They can capture more visual information and offer the advantage of redundancy, which helps improve robustness for localization and mapping. As a result, multi-camera sensing has attracted much research interest in recent times leading to novel SLAM solutions \cite{sola2008fusing}\cite{Kaess2010}\cite{kuo2020redesigning}, datasets\cite{helmberger2021hilti}, and implementations on robotic platforms\cite{Heng2018}.

Many earlier works on multi-camera SLAM either treat cameras independently\cite{sola2008fusing}\cite{urban2016multicol}\cite{zhang2022mmo} or assume a specific setup\cite{Carrera2011}\cite{Tribou2015}\cite{heng2015self}\cite{seok2019rovo}, and do not exploit the camera arrangement to the fullest. In this paper, we examine the much more general case of multiple overlapping and
non-overlapping cameras. 
 We use a generalized camera model\cite{pless2003using} to represent the multi-camera system as a collection of unconstrained rays without making any assumptions about a particular geometry. We build upon the ideas presented in \cite{kuo2020redesigning} but differ significantly in terms of modeling the camera system, making our approach extensible to multiple overlapping views beyond stereo pairs. Another challenge with multiple cameras is to effectively and efficiently utilize the increased amount of information provided by the sensors. We compute overlapping regions among cameras in the rig and extract 3D features via cross-matching. This leverages the camera configuration to fuse multi-view data, avoids duplicating features, and keeps the computational cost in check. 

Realizing practical mobile robotics applications requires designing sensor systems along with algorithmic development. The choice of cameras and the system configuration of the sensing platform impacts the SLAM outcome\cite{zhang2016benefit}\cite{chappellet2021benchmarking}. While there have been significant efforts toward developing SLAM algorithms for different sensors, there has been a lack of attention to the system design aspects of the sensing platform. To address this gap, we identify the number of cameras and the overlap between their FoVs as two key design parameters of the system configuration that impact the information gathered by the cameras and, thus, SLAM estimates. We evaluate the localization accuracy and robustness of our SLAM system on several real-world datasets collected using our custom-built multi-camera rig. 
The major contributions of this paper include
\begin{itemize}
    \item A real-time generic visual SLAM framework that models a multi-camera system as a generalized camera and computes novel 3D intra-match features to fully utilize the overlapping views with open source code\footnote{ \label{note1} \scriptsize Code is available at \url{https://github.com/neufieldrobotics/MultiCamSLAM}}.
    \item Evaluation of the SLAM method in terms of tracking accuracy, robustness, and computational constraints.
    \item A systematic empirical evaluation of different multi-camera system configurations through experiments on real-world datasets in terms of the number of cameras and overlapping vs non-overlapping camera geometries.
    \item A collection of new real-world datasets\footnote{ \scriptsize Data available at \url{https://tinyurl.com/mwfkrj8k}} used to evaluate the proposed SLAM system. These datasets allow us to explore more realistic settings with longer runs and far more varying dynamic content than extant SLAM datasets. 
\end{itemize}

%% file: sections/relatedwork.tex
\section{Related Work}
There is a large body of literature on visual SLAM that focuses on monocular, stereo or RGBD cameras and exploits their intrinsic and extrinsic properties. For instance, stereo and RGBD cameras are used to obtain metric SLAM estimates
\cite{engel2015large}\cite{kerl2013robust} in contrast to monocular SLAM\cite{davison2003real} with bearing-only image data. Within the monocular realm, wide field-of-view (FoV) and omnidirectional cameras have been used for SLAM
\cite{matsuki2018omnidirectional}. There are many state-of-the-art SLAM pipelines that work off the shelf
\cite{campos2021orb}\cite{Forster2017}.

Multi-camera SLAM has recently gained attention due to its robustness and superior perception capabilities. One of the first multi-camera SLAM frameworks was discussed by Sola et al\cite{sola2008fusing}, where data from multiple independent monocular cameras are fused via filtering. Kaess et al\cite{Kaess2010} propose a probabilistic approach to iteratively solve data association and SLAM for a circular 8-camera rig. They show that allowing cameras to face different directions yields better constraints and helps with loop closures irrespective of robot orientation. MultiColSLAM \cite{urban2016multicol} extends ORBSLAM to a rigidly coupled fish-eye multi-camera system by modeling the camera cluster as a generalized camera \cite{pless2003using}. Zhang et al\cite{zhang2022mmo} present a SLAM system that also calibrates extrinsic parameters using wheel odometry. These works mainly target non-overlapping camera setups that provide maximal coverage of the surroundings and do not exploit the overlap between cameras. Other similar contributions include \cite{Tribou2015}\cite{Carrera2011}\cite{Houben2016}. 

A few works employ stereo pairs to obtain 3D features via stereo matching between overlapping areas and recover metric scale. Heng et al\cite{heng2015self} use four fish-eye cameras arranged as two stereo pairs facing forward and backwards to give a 360$^{\circ}$ view. ROVO\cite{seok2019rovo} uses the same camera configuration but applies a hybrid projection model to minimize distortion and maximize feature matching. In \cite{Jaekel2020}, a multi-stereo pipeline is presented with a new 1-point RANSAC for joint outlier rejection. In \cite{Liu2018} and its follow-up work\cite{Heng2018},  multiple stereo pairs improve the feature space for tracking while driving at night from several real-world experiments. 
In\cite{kaveti2021towards}, a linear front-facing camera array is used to perform Light Field based static background reconstruction to handle dynamic objects. Li et al \cite{li2021robust}, perform robust initialization and extrinsic calibration in multi-camera systems with limited view overlaps. Zhang et al\cite{zhang2021balancing} use submatrix feature selection to reduce compute and track features across stereo and mono cameras. However, these methods assume a priori knowledge about the camera configuration.
Kuo et al\cite{kuo2020redesigning}, present an adaptive SLAM design to accommodate arbitrary camera systems. However, it is not clear if the system can handle overlapping configurations beyond stereo since the method focuses on finding stereo pairs and all the experiments were evaluated on stereo pairs alone. 

Studying the impact of camera system design will enable us to build robust multi-camera systems for mobile robots. Nevertheless, the research in this direction is still in a nascent stage. Zhang et al\cite{zhang2016benefit} discuss the impact of FoV of a monocular camera on visual odometry in urban environments. Kevin et al\cite{chappellet2021benchmarking} benchmark commercially available monocular, stereo and RGB-D sensors by evaluating them on the OpenVSLAM framework. In \cite{Tribou2015}, the authors evaluate MCPTAM on overlapping and non-overlapping camera clusters and show that overlapping cameras recover accurate scale and cause less drift. We add to these efforts by evaluating the impact of the number of cameras and their extrinsic arrangement on SLAM. 

%% file: sections/slam_framework.tex
\begin{figure}[ht!]
\vspace{2mm}
\centering
\captionsetup{font={footnotesize}, labelfont={bf,sf}}
\begin{tabular}{c}
\subfloat[]{
\includegraphics[clip, trim={0cm 0cm 0cm 0cm},width=\linewidth]{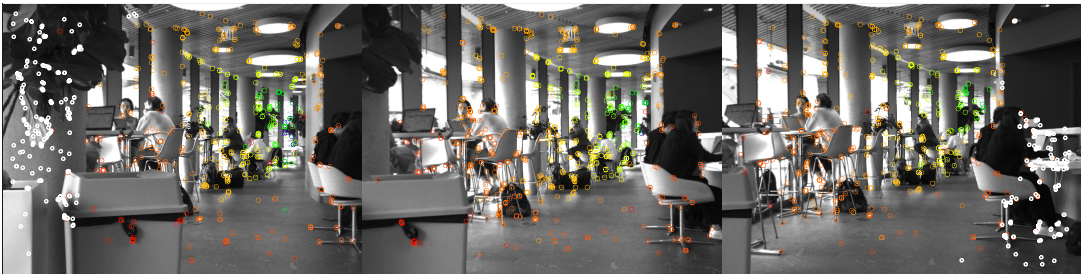}} \\
\subfloat[]{\includegraphics[clip, trim={0cm 0cm 0cm 0cm},width=\linewidth]{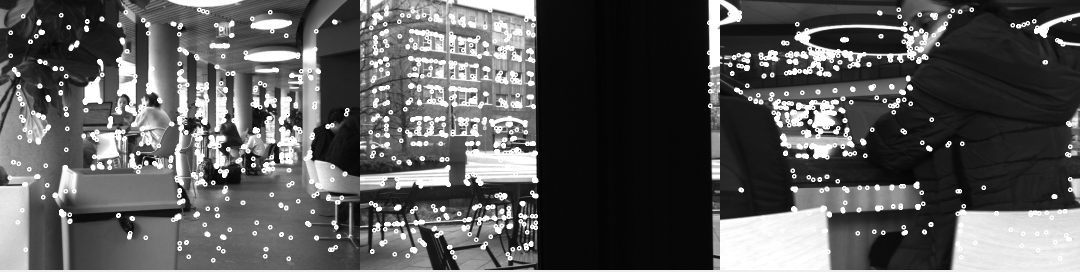}}
\end{tabular}
\caption{Two sample multi-camera frames with extracted features on (a) front-facing cameras in overlapping (OV) and, (b) non-overlapping (N-OV) setups for the same scene. The multi-view metric features are colored based on their distance to the camera system. The white points are mono features that do not have any 3D information. Notice that N-OV setup has only mono features, whereas the OV setup has both multi-view and mono features distributed across OV and N-OV regions in the images.}
\label{fig:intra_matches}
\vspace{-2mm}
\end{figure}

\section{Problem Setup}
The primary goal of our VSLAM formulation is to develop a unified framework independent of the camera system configuration, which is lightweight and runs in real-time on mobile robots. A block diagram of the generic visual SLAM pipeline is shown in \cref{fig:intro_pic}.

\textbf{Definitions:} Let the number of cameras in the sensor rig be $N_c >=1$. Each component camera $c_p$ is related to the body frame of the multi-camera rig by a rigid body transformation ${}^{B}T_{p}=\{{}^{B}R_{p},{}^{B}t_{p}\}$ obtained via extrinsic calibration parameters. The intrinsic parameters of the component cameras are denoted as $K_p$. 
Estimated keyframe poses are represented as $x_n \in SE(3)$, $n \in {1,..., N}$ and correspond to the rig body frame with respect to the world ${}^{w}T_B$. The landmarks are represented as $l_m \in R^3 $ with $m \in {1,..., M}$.  The visual measurements of the landmarks in the images are denoted as $z_k \in R^2$ with $k \in {1,..., K}$, the data association that relates each measurement $z_k$ with a keyframe, component camera, and a landmark is given by $(x_{n_k}, c_{p_k}, l_{m_k})$. We assume the body frame of the rig to align with one of the camera coordinate frames for simplicity.

\textbf{Camera Configuration:} An arbitrary camera configuration is the most generic definition, which encompasses all the ways a set of cameras can be rigidly arranged on a robotics platform. For example, we can use monocular or stereo setups, or cameras lying on a circular ring with the possibility of overlapping or non-overlapping FoV. We distinguish different camera configurations based on the number of cameras ($N_c$) and the FoV overlap among them and benchmark their influence on SLAM. For this study, for the purposes of clarity, we have scoped our work solely on either overlapping (OV) or non-overlapping (N-OV) (shown in \cref{fig:intro_pic}a) scenarios, even though the methodology is generally applicable to mixing both overlapping and non-overlapping camera rigs. Irrespective of the configuration, the multi-camera rig is considered a single generalized camera that captures a collection of rays passing through multiple pinholes which will be detailed in section IV. 
\section{The Front-end}
The front end of a SLAM system aims to estimate the pose of the robot and the landmarks observed at each time step. This section discusses the critical aspects of feature extraction and representation, initialization, tracking, and mapping modules that enable us to seamlessly work with arbitrary multi-camera systems.
\subsection{Feature Extraction}
We perform sparse feature tracking using two types of features - \textbf{Multi-view features} and \textbf{Mono features} [Ref. \cref{fig:intra_matches}]. Depending on their placement, the cameras comprising the camera rig can have overlapping fields of view. We leverage the overlapping imagery to compute strong metric features. Instead of using the features extracted from component cameras independently, we associate overlapping image regions to group features that belong to a specific 3D point in the scene. This differs from most existing camera systems which do not utilize overlap among cameras except to compute stereo. Our multi-view features let us accurately represent the scene with fewer features and avoid creating redundant landmarks.\\
\textbf{Determining Overlap:} This is the first step required to compute multi-view features efficiently. For each component camera $c_i$, we find the common image regions in the set of cameras {$\mathcal{C}$} that share a common FoV with $c_i$. Starting with a camera pair ($c_i, c_j$), we first divide the image of $c_i$ into a 2D grid. For each grid cell $g_k$, the epipolar line $e_{ij}^k$ corresponding to its center $g_k^x, g_k^y$ is computed using extrinsic calibration between the camera pair as shown below. 
\begin{align}
    F_{ij} = K_j[{}^{j}t_i]_\times {}^{j}R_i K_i^{-1} \\
    e_{ij}^k = F_{ij} [g_k^x, g_k^y]
\end{align}
Next, we determine the grid cells in the image of $c_j$ that the epipolar line passes through and save them as potential cells to find correspondences. 
If the line does not pass through the image plane, we know that the camera pair is non-overlapping. This computation is done just once during initialization in the beginning of and then used throughout the run while executing the SLAM framework.\\
\textbf{Multi-view Features:} These are essentially cross-matches between cameras. First, we extract multi-scale ORB features in all images and assign them to the 2D grid. This process is parallelized for speed. We then iteratively compute feature correspondences across cameras resulting in a total of $^{N_c}\mathcal{C}_2$ combinations for $N_c$ cameras. For a camera pair ($c_i, c_j$), instead of matching every feature in $c_i$ to every feature in $c_j$, we match features cell-wise based on the overlap to reduce computation. For a set of features, $\mathcal{F}1$, belonging to a cell in the image of $c_i$ we obtain the set of features $\mathcal{F}2$ lying in the cells corresponding to the potential matches obtained from the overlap determination step. We then perform brute-force matching between the sets $\mathcal{F}1$ and $\mathcal{F}2$. Each match is then passed through the epipolar constraint which checks if the corresponding feature in the second view is within a certain distance from the epipolar line. 
A set of matches $\mathcal{M}$ is created from the first pair of cameras. For the subsequent image pairs, if correspondence is found between two unmatched features, a new match is added to the set of matches $\mathcal{M}$. If a match is found for an already matched feature, the new feature is added to the existing match. 

Each multi-view match represents a 3D feature in the scene. It consists of a bundle of rays given by a set of 2D keypoints, which are projections of the 3D feature in component cameras, the 3D coordinate obtained from triangulation, and a representative descriptor computed to facilitate feature matching for tracking. Note that the 3D point does not need to be observed in all the cameras. The observability depends on the overlap and scene structure. 

\textbf{Mono Features} In the case of either a monocular camera or non-overlapping camera configurations, there are no multi-view matches. Even within overlapping camera configurations, there might be some non-overlapping regions depending on the structure of the 3D scene. We have mono features with a single 2D keypoint and its descriptor to represent the non-overlapping areas.
\begin{figure}[t]
    \vspace{2mm}
    \centering
    \captionsetup{font={footnotesize},labelfont={bf,sf}}
    \includegraphics[width=\columnwidth]{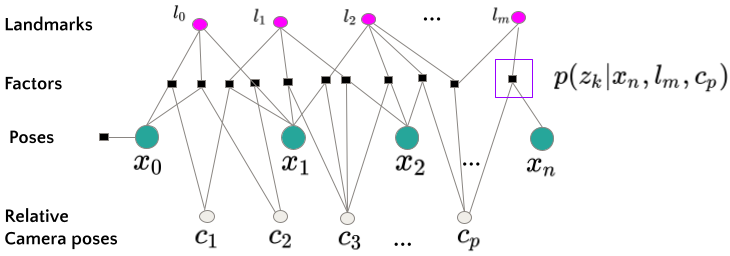}
    \caption[Multi-camera back-end factor graph]{Factor graph of the multi-camera back-end with poses $X_i$, landmarks $l_j$ and the relative camera poses $C_p$ as the variables to be optimized. The black square factor nodes represent constraints on the variables.}
    \label{fig:mc_backend}
    \vspace{-2mm}
\end{figure}

\subsection{Initialization}
This step creates the initial set of landmarks used to track subsequent frames. we perform initialization based on the underlying camera configuration. After extracting features, if the number of metric multi-view features is greater than a certain threshold, we utilize them as the initial map. Otherwise, we must choose two initial frames and compute the relative pose between them. We use a generalized camera model which allows non-central projection suitable to represent multi-camera systems\cite{pless2003using}. A key point expressed as a pixel, $u$, in the image of camera $c_p$, is represented as a Plucker line $L=[q\;, q']$ which is a 6-vector made of line direction $q$ and moment $q'$. 
\begin{equation}
    \hat{u} = K_{p}^{-1} u, \;\; q = {}^{B}R_{p}\hat{u}, \;\;  q^{'} = {}^{B}t_{p} \times q 
\end{equation}
we determine correspondences between two frames and solve the generalized epipolar constraint to obtain relative pose
\begin{equation}
     \begin{pmatrix}
      q_2  \\
      q_2^{'}  
     \end{pmatrix}^T \begin{pmatrix}
      E & R  \\
      R & 0 
      \end{pmatrix} \begin{pmatrix}
                     q_1  \\
                     q_1^{'}  
                     \end{pmatrix} = 0 \label{eq:gec3}
\end{equation}
where $[q_1 \;, q_1^{'}]$ and $[q_2 \;, q_2^{'}]$ are the Plucker rays of matched features, $E=[t]_{\times}R$ is the essential matrix, and $R$ and $t$ are the rotation and translation between the two generalized camera frames.  In the case of a monocular camera, the Plucker ray corresponding to the keypoints is $[\hat{u} \;, 0]$ from eq(3). Thus the generalized epipolar constraint is naturally reduced to the classical epipolar constraint.
We apply the 5-point algorithm for a monocular camera and the 17-point algorithm\cite{li2008linear} for a multi-camera rig with RANSAC for relative pose estimation. We make sure that there is sufficient parallax between the two frames and triangulate the correspondences which act as our initial landmarks, and follow this with non-linear optimization to obtain our final landmark estimates.
\subsection{Tracking and Mapping}
After initialization, every incoming frame is tracked with respect to the last keyframe. We compute inter-frame correspondences between the last keyframe and the current frame via bag-of-words matching. Since multi-view features contain more than one descriptor from different component cameras we use the median of descriptors for matching. If enough 3D-2D matches are found between the map points ${}^{w}P_i$ in the last keyframe and observations $z_k$ in the current frame we obtain the Plucker coordinates of $z_k$, $[q_k\;,q_k^{'}]$ using eq (3) and estimate the absolute pose of the current frame $[{}^{B}R_w \;\; {}^{B}t_w]$ using generalized PnP\cite{kneip2014upnp} by solving a set of constraints of the form 
\begin{equation}
    \lambda q_i + q_i \times q_i^{'} = {}^{B}R_w {}^{w}P_i + {}^{B}t_w 
\end{equation}
If the estimated pose indicates significant motion since the last keyframe we further localize the current frame with respect to the local map in a manner similar to ORBSLAM \cite{campos2021orb}. We find the set of neighboring keyframes $\mathcal{K}$ shared by the initially tracked landmarks. We then compute new matches between the landmarks tracked in $\mathcal{K}$ and the current frame. This enables us to gain local map support and helps in finding stable landmarks in the presence of occluding dynamic objects. Finally, the current frame is inserted as a keyframe if the ratio of the tracked landmarks since the last keyframe is less than a certain threshold. When a new keyframe decision is made, the observations are added to the existing landmarks and the new inter-frame matches corresponding to the non-map points are triangulated to create new map points.
\input{tables/data_desc}
\section{The Back-end}
The back end corresponds to the optimization framework that refines the initial estimates of keyframe poses $X$ and landmarks $L$ by maximizing the posterior probability on the variables given observations $Z = \{z_k\}_{k=1}^{K}$\cite{dellaert2017factor}. In a general multi-camera system an observation not only depends on the pose of the rig $X$ and the landmark $L$, but also on the component cameras $C$ it is seen in. The Maximum a Posteriori (MAP)  problem is given by 
\begin{align}
   X^*, L^*, C^* &= \underset{X, L, C} \argmax \; P(X,L,C\; | Z) \nonumber \\
        &\propto \underset{X, L,C}\argmax \; P(Z\; | X,L,C) \; P(X, L,C) \nonumber \\
        &\propto \underset{X, L,C}\argmax \; P(x_0) \prod_{k=1}^{K} P(z_k | x_{n_k},l_{m_k}, c_{p_k})  
\end{align}
where $P(Z | X,L,C)$ is the likelihood of the observations which factorizes into a product of individual probabilities due to i.i.d assumption. $P(x_0)$ is the prior probability over the initial robot pose. The factor graph representation in \cref{fig:mc_backend} shows the individual probability constraints(factors) among variables. 
Assuming normally distributed zero-mean noise for the observations $z_k$ and modeling prior also as a gaussian 
eq(6) takes the least-squares minimization form
\begin{align}
    X^*, L^*, C^* \triangleq \underset{X, L, C} \argmin \sum_{k=1}^{K} \| {h_k(x_{n_k}, l_{m_k}, c_{p_k}) - z_k } \|^2_{\Sigma_k} \nonumber\\
    - \log p(x_0) \nonumber 
\end{align}
where the measurement function $h_k(x_{i_k}, l_{j_k}, c_{p_k})$ maps a landmark to a predicted observation $\hat{z_k}$ through a series of transformations. First, the pose of the body $x_{n_k} = {}^{w}T_{B}^n$ and the relative pose of the component camera $ c_{p_k} = {}^{B}T_{p_k}$ are used to get the pose of the camera in world frame via SE(3) composition ${}^{w}T_c = {}^{w}T_{B}^n. {}^{B}T_c$. The 3D landmark $l_{m_k} = P^w$ is transformed from world frame into the camera coordinate frame $P^c = f(T_c^w, P^w) = (R_c^w)^T(P^w - t_c^w)$. Finally, $P^c$ is projected to 2D image coordinates using the intrinsic matrix $\hat{z_k} = K_cP^c$.
This formulation conveniently models multi-view features. It gives the back end the flexibility to work with different camera configurations and optimize the extrinsic calibration parameters of the component cameras $C$ along with estimating the trajectory and landmarks.

%% file: tables/data_desc.tex
\begin{table*}[t!]
\vspace{2mm}
\centering
\scriptsize
\captionsetup{font=footnotesize,labelfont={bf,sf}}
\resizebox{0.9\textwidth}{!}{%
\begin{tabular}{|lllllll|}
\hline
\multicolumn{7}{|c|}{\textbf{Datasets}} \\ \hline
\multicolumn{1}{|l|}{\textbf{Label}} &
  \multicolumn{1}{l|}{\textbf{Location}} &
  \multicolumn{1}{l|}{\textbf{Total Frames}} &
  \multicolumn{1}{l|}{\textbf{Length (m)}} &
  \multicolumn{1}{l|}{\textbf{Groundtruth}} &
  \multicolumn{1}{l|}{\textbf{Attributes}} &
  \textbf{Loop} \\ \hline
\multicolumn{1}{|l|}{ISEC\_Lab1} &
  \multicolumn{1}{l|}{indoor Lab} &
  \multicolumn{1}{l|}{9130} &
  \multicolumn{1}{l|}{152} &
  \multicolumn{1}{l|}{optitrack} &
  \multicolumn{1}{p{4cm}|}{feature less areas , reflective surfaces} &
  Yes \\ \hline
\multicolumn{1}{|l|}{ISEC\_Ground1} &
  \multicolumn{1}{l|}{indoor lobby} &
  \multicolumn{1}{l|}{4481} &
  \multicolumn{1}{l|}{90} &
  \multicolumn{1}{l|}{target based} &
  \multicolumn{1}{p{4cm}|}{reflective surfaces, minimal dynamic content, day light} &
  Yes \\ \hline
\multicolumn{1}{|l|}{ISEC\_Ground2} &
  \multicolumn{1}{l|}{indoor lobby} &
  \multicolumn{1}{l|}{5408} &
  \multicolumn{1}{l|}{90} &
  \multicolumn{1}{l|}{target based} &
  \multicolumn{1}{p{4cm}|}{reflective surfaces, minimal dynamic content, night} &
  Yes \\ \hline
\multicolumn{1}{|l|}{ISEC\_Ground3} &
  \multicolumn{1}{l|}{indoor lobby} &
  \multicolumn{1}{l|}{6001} &
  \multicolumn{1}{l|}{120} &
  \multicolumn{1}{l|}{target based} &
  \multicolumn{1}{p{4cm}|}{reflective surfaces,high dynamic content, day light} &
  Yes \\ \hline
\multicolumn{1}{|l|}{Falmouth} &
  \multicolumn{1}{l|}{outdoor offroad} &
  \multicolumn{1}{l|}{9156} &
  \multicolumn{1}{l|}{2800} &
  \multicolumn{1}{l|}{target based + gps} &
  \multicolumn{1}{p{4cm}|}{fast motion, outdoor, foliage, minimal dynamic content, large area} &
  Yes \\ \hline
\multicolumn{1}{|l|}{Curry\_center} &
  \multicolumn{1}{l|}{campus} &
  \multicolumn{1}{l|}{28526} &
  \multicolumn{1}{l|}{590} &
  \multicolumn{1}{l|}{target based} &
  \multicolumn{1}{p{4cm}|}{Urban,outdoor,dynamiccontent,largearea} &
  Yes \\ \hline
\multicolumn{1}{|l|}{ISEC\_floor5} &
  \multicolumn{1}{l|}{campus} &
  \multicolumn{1}{l|}{4300} &
  \multicolumn{1}{l|}{187} &
  \multicolumn{1}{l|}{target based} &
  \multicolumn{1}{p{4cm}|}{Urban,indoor,repeated structures} &
  Yes \\ \hline
\end{tabular}%
}
\caption{This table describes the attributes of the dataset used to evaluate the proposed SLAM framework. We collected six sequences with trajectory lengths ranging between 90 m - 2800 m in a wide variety of indoor and outdoor environments.}
\label{tab:dataset_desc}
\vspace{-3mm}
\end{table*}

%% file: sections/experimental_setup.tex
\section{Experimental Setup}
In this section, we describe the hardware setup, calibration, and datasets collected for the experimental evaluation. 
\input{tables/mcslam_allconfigs_accuracy}
\subsection{Hardware Setup and Camera Calibration}
\begin{figure}[h!]
\vspace{-2mm}
    \centering
     \captionsetup{font={footnotesize},labelfont={bf,sf}}
    \includegraphics[width=\columnwidth, height=110pt]{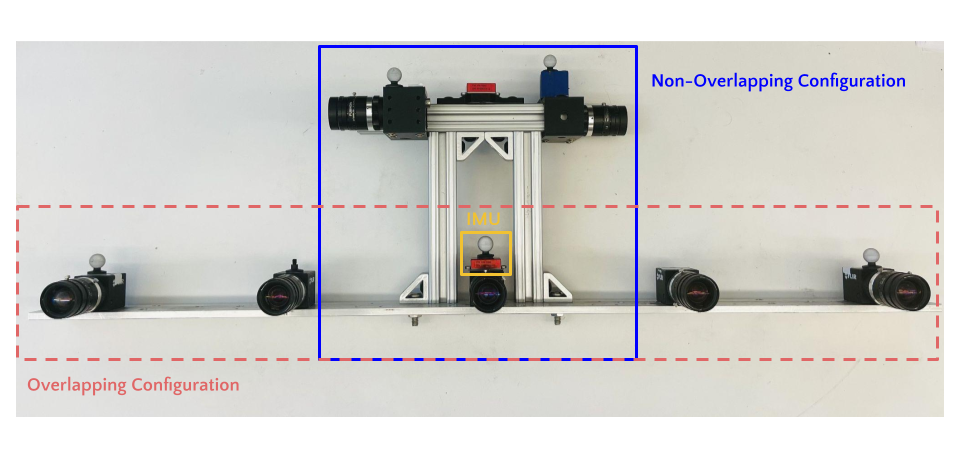}
    \caption{The custom-built multi-camera rig used to collect data for evaluating the SLAM pipeline. The figure shows overlapping and non-overlapping configurations and the IMU that was mounted on the rig. The IMU is used to compute the baseline between two consecutive cameras is 165mm.}
    \label{fig:camera_rig}
    \vspace{-2mm}
\end{figure}

We use a rigid multi-camera rig consisting of 7 cameras with 5 cameras facing forward and 2 sideways and an inertial measurement unit (IMU) as shown in figure \cref{fig:camera_rig}. The cameras are arranged to accommodate configurations with both overlapping(OV) and non-overlapping(N-OV) FoV as shown in \cref{fig:camera_rig}. The front-facing cameras(red dashed box) are used to run experiments for mono, stereo, and OV multi-camera setups. The center front-facing and side-facing cameras(blue box) are used as N-OV multi-camera setup. We use FLIR BlackFly 1.3 MP color cameras with a resolution of 720 x 540, 57$^\circ$ FoV and a Vectornav IMU running at 200 HZ. All the cameras are hardware synchronized at 20 fps. 
We use Kalibr \cite{furgale2013unified} to obtain the intrinsic and extrinsic parameters of the cameras with overlapping FoV. For non-overlapping cameras, target-based calibration does not work as we need the cameras to observe the target to solve for the relative transformation. Instead, we perform imu-camera calibration for each of the non-overlapping cameras and chain the inter-camera transformations together. This gives a good initial estimate of extrinsic parameters, which are refined during optimization as described in section V.
\input{tables/accuracy_stereo}
\input{tables/ablation_mcslam}

\subsection{Datasets}
We collected a set of six indoor and outdoor sequences. The multi-camera rig along with a DELL XPS laptop with 32GB RAM was mounted on a Clearpath Ridgeback robotic platform and driven across Northeastern University's campus for data collection. These sequences include several challenging scenes consisting of narrow corridors, featureless spaces, jerky motions, sudden turns, and dynamic objects which are commonly encountered in urban environments. One of the sequences was collected with the NUANCE Autonomous car in an offroad environment. Ground truth is obtained from GPS for the offroad sequence and using Optitrack (accurate up to a millimeter) for the indoor sequence in the lab area. In cases where Optitrack could not be used, visual tags are used as ground truth to compute drift. The details of the dataset including location, length of the trajectory, and ground truth are consolidated in \cref{tab:dataset_desc}.   

%% file: tables/mcslam_allconfigs_accuracy.tex

\begin{table*}[]
\vspace{2mm}
\captionsetup{font=footnotesize,labelfont={bf,sf}}
\resizebox{\textwidth}{!}{%
\begin{tabular}{|l|ll|llllllll|llllll|}
\hline
\multirow{3}{*}{} &
  \multicolumn{2}{c|}{\multirow{3}{*}{\textbf{\begin{tabular}[c]{@{}c@{}}Trajectory\\  length (m)\end{tabular}}}} &
  \multicolumn{8}{c|}{\textbf{OV}} &
  \multicolumn{6}{c|}{\textbf{N-OV}} \\ \cline{4-17} 
 &
  \multicolumn{2}{c|}{} &
  \multicolumn{2}{c|}{2cams} &
  \multicolumn{2}{c|}{3cams} &
  \multicolumn{2}{c|}{4cams} &
  \multicolumn{2}{c|}{5cams} &
  \multicolumn{2}{c|}{1cam} &
  \multicolumn{2}{c|}{2cams} &
  \multicolumn{2}{c|}{3cams} \\ \cline{4-17} 
 &
  \multicolumn{2}{c|}{} &
  \begin{tabular}[c]{@{}l@{}}ATE\\  (m)\end{tabular} &
  \multicolumn{1}{l|}{scale} &
  \begin{tabular}[c]{@{}l@{}}ATE\\  (m)\end{tabular} &
  \multicolumn{1}{l|}{scale} &
  \begin{tabular}[c]{@{}l@{}}ATE \\ (m)\end{tabular} &
  \multicolumn{1}{l|}{scale} &
  \begin{tabular}[c]{@{}l@{}}ATE\\  (m)\end{tabular} &
  scale &
  \begin{tabular}[c]{@{}l@{}}ATE\\  (m)\end{tabular} &
  \multicolumn{1}{l|}{scale} &
  \begin{tabular}[c]{@{}l@{}}ATE\\  (m)\end{tabular} &
  \multicolumn{1}{l|}{scale} &
  \begin{tabular}[c]{@{}l@{}}ATE\\  (m)\end{tabular} &
  scale \\ \hline
\multirow{3}{*}{\textbf{MCSLAM}} &
  \multicolumn{1}{l|}{\textbf{Plus}} &
  11.754 &
  0.0226 &
  \multicolumn{1}{l|}{1.007} &
  0.022 &
  \multicolumn{1}{l|}{1.01} &
  0.020 &
  \multicolumn{1}{l|}{1.01} &
  \textbf{0.0194} &
  \textbf{1.01} &
  0.027 &
  \multicolumn{1}{l|}{3.52} &
  0.023 &
  \multicolumn{1}{l|}{0.15} &
  0.022 &
  0.204 \\
 &
  \multicolumn{1}{l|}{\textbf{Square}} &
  9.898 &
  0.011 &
  \multicolumn{1}{l|}{1.008} &
  0.01 &
  \multicolumn{1}{l|}{1.007} &
  0.009 &
  \multicolumn{1}{l|}{1.006} &
  \textbf{0.009} &
  \textbf{1.008} &
  0.011 &
  \multicolumn{1}{l|}{2.01} &
  0.018 &
  \multicolumn{1}{l|}{0.29} &
  0.0107 &
  0.282 \\
 &
  \multicolumn{1}{l|}{\textbf{Circle}} &
  11.166 &
  0.039 &
  \multicolumn{1}{l|}{0.806} &
  0.042 &
  \multicolumn{1}{l|}{0.807} &
  \textbf{0.040} &
  \multicolumn{1}{l|}{\textbf{0.95}} &
  0.043 &
  0.98 &
  0.412 &
  \multicolumn{1}{l|}{4.42} &
  0.23 &
  \multicolumn{1}{l|}{0.425} &
  0.086 &
  0.117 \\ \hline
\multirow{3}{*}{\textbf{\begin{tabular}[c]{@{}l@{}}MultiCol\\ SLAM\end{tabular}}} &
  \multicolumn{1}{l|}{\textbf{Plus}} &
  11.754 &
  0.03 &
  \multicolumn{1}{l|}{1.39} &
  0.16 &
  \multicolumn{1}{l|}{1.14} &
  0.106 &
  \multicolumn{1}{l|}{1.248} &
  0.041 &
  1.242 &
  0.021 &
  \multicolumn{1}{l|}{2.557} &
  0.161 &
  \multicolumn{1}{l|}{1.431} &
  0.23 &
  0.684 \\
 &
  \multicolumn{1}{l|}{\textbf{Square}} &
  9.898 &
  0.022 &
  \multicolumn{1}{l|}{1.522} &
  0.068 &
  \multicolumn{1}{l|}{0.93} &
  0.142 &
  \multicolumn{1}{l|}{1.408} &
  0.102 &
  1.646 &
  0.571 &
  \multicolumn{1}{l|}{1.41} &
  0.54 &
  \multicolumn{1}{l|}{0.959} &
  0.695 &
  0.578 \\
 &
  \multicolumn{1}{l|}{\textbf{Circle}} &
  11.166 &
  - &
  \multicolumn{1}{l|}{-} &
  - &
  \multicolumn{1}{l|}{-} &
  - &
  \multicolumn{1}{l|}{-} &
  - &
  - &
  - &
  \multicolumn{1}{l|}{-} &
  - &
  \multicolumn{1}{l|}{-} &
  - &
  - \\ \hline
\end{tabular}%
}

\caption{Absolute trajectory errors and scale correction across several overlapping (OV) and non-overlapping(N-OV) configurations with respect to MultiColSLAM, the baseline multi-camera SLAM method. For OV setup, we choose cameras from the front-facing array, starting with a stereo setup and adding one camera for each experiment. For the N-OV setup, we choose the 1, 2, and three cameras, respectively, from the set shown in the enclosed blue box in {\cref{fig:camera_rig}} . We outperform the baseline in all sequences.  The empty entries correspond to tracking loss. We present the results on three additional simple trajectories with optitrack ground truth, as MultiColSLAM did not work on any datasets shown in   fig. {\ref{tab:dataset_desc}}, which are significantly large and challenging.}

\label{tab:mcslam_compare}
\vspace{-3mm}
\end{table*}

%% file: tables/accuracy_stereo.tex
\begin{table}[]
\captionsetup{font=footnotesize,labelfont={bf,sf}}
\resizebox{\columnwidth}{!}{%
\begin{tabular}{|l|ll|ll|ll|}
\hline
\multirow{2}{*}{\textbf{\begin{tabular}[c]{@{}l@{}}Sequence/\\  Algorithm\end{tabular}}} &
  \multicolumn{2}{c|}{\textbf{MCSLAM}} &
  \multicolumn{2}{c|}{\textbf{ORBSLAM3}} &
  \multicolumn{2}{c|}{\textbf{SVO Pro}} \\ \cline{2-7} 
 &
  \multicolumn{1}{l|}{\textbf{ATE (m)}} &
  \textbf{ATE (\%)} &
  \multicolumn{1}{l|}{\textbf{ATE (m)}} &
  ATE (\%) &
  \multicolumn{1}{l|}{\textbf{ATE (m)}} &
  \textbf{ATE (\%)} \\ \hline
\textbf{ISECLab1} &
  \multicolumn{1}{l|}{2.65} &
  1.74 &
  \multicolumn{1}{l|}{2.9} &
  1.91 &
  \multicolumn{1}{l|}{\textbf{2.435}} &
  \textbf{1.6} \\
\textbf{ISEC Ground1} &
  \multicolumn{1}{l|}{0.99} &
  1.07 &
  \multicolumn{1}{l|}{\textbf{0.85}} &
  \textbf{0.8} &
  \multicolumn{1}{l|}{0.901} &
  1.01 \\
\textbf{ISEC Ground2} &
  \multicolumn{1}{l|}{\textbf{2.41}} &
  \textbf{2.51} &
  \multicolumn{1}{l|}{2.98} &
  3.1 &
  \multicolumn{1}{l|}{2.698} &
  2.99 \\
\textbf{Curry Center} &
  \multicolumn{1}{l|}{\textbf{1.18}} &
  \textbf{1.5} &
  \multicolumn{1}{l|}{4.84} &
  6.3 &
  \multicolumn{1}{l|}{41.236} &
  11.87 \\
\textbf{Falmouth} &
  \multicolumn{1}{l|}{210.02} &
  8.05 &
  \multicolumn{1}{l|}{\textbf{49.708}} &
  \textbf{1.91} &
  \multicolumn{1}{l|}{61.38} &
  2.19 \\
\textbf{ISEC floor 5} &
  \multicolumn{1}{l|}{\textbf{0.501}} &
  \textbf{0.27} &
  \multicolumn{1}{l|}{0.516} &
  0.28 &
  \multicolumn{1}{l|}{0.626} &
  0.33
   \\ \hline
\end{tabular}%
}
\caption{Absolute translation error(ATE) of the proposed method and two SOTA stereo SLAM systems SVO and ORBSLAM3 with respect to the ground truth for different sequences. The table shows that the stereo configurations for all systems are roughly comparable.}
\label{tab:slam_accuracy_stereo}
\vspace{-2mm}
\end{table}

%% file: tables/ablation_mcslam.tex
\begin{table*}[]
\vspace{3mm}
\captionsetup{font=footnotesize,labelfont={bf,sf}}
\resizebox{\textwidth}{!}{%
\captionsetup{font=footnotesize,labelfont={bf,sf}}
\resizebox{\textwidth}{!}{%
\begin{tabular}{|l|
>{\columncolor[HTML]{FFFFFF}}l 
>{\columncolor[HTML]{FFFFFF}}l 
>{\columncolor[HTML]{FFFFFF}}l 
>{\columncolor[HTML]{FFFFFF}}l 
>{\columncolor[HTML]{FFFFFF}}l 
>{\columncolor[HTML]{FFFFFF}}l 
>{\columncolor[HTML]{FFFFFF}}l 
>{\columncolor[HTML]{FFFFFF}}l |
>{\columncolor[HTML]{FFFFFF}}l 
>{\columncolor[HTML]{FFFFFF}}l 
>{\columncolor[HTML]{FFFFFF}}l 
>{\columncolor[HTML]{FFFFFF}}l |}
\hline
& \multicolumn{8}{c|}{\cellcolor[HTML]{FFFFFF}\textbf{Overlapping configuration}}                             & \multicolumn{4}{c|}{\cellcolor[HTML]{FFFFFF}\textbf{Non-Overlapping configuration}}                         \\ \cline{2-13} 
& \multicolumn{2}{c|}{\cellcolor[HTML]{FFFFFF}\textbf{2cams (stereo)}}                                                & \multicolumn{2}{c|}{\cellcolor[HTML]{FFFFFF}\textbf{3cams}}                                                 & \multicolumn{2}{c|}{\cellcolor[HTML]{FFFFFF}\textbf{4cams}}                                                 & \multicolumn{2}{c|}{\cellcolor[HTML]{FFFFFF}\textbf{5cams}}
& \multicolumn{2}{c|}{\cellcolor[HTML]{FFFFFF}\textbf{ 2cams}}                                                 & \multicolumn{2}{c|}{\cellcolor[HTML]{FFFFFF}\textbf{3cams}}                       
\\ \cline{2-13} 
\multirow{-3}{*}{}     
& \multicolumn{1}{l|}{\cellcolor[HTML]{FFFFFF}\textbf{ATE (m)}} 
& \multicolumn{1}{c|}{\cellcolor[HTML]{FFFFFF}\textbf{ATE (\%)}} 
& \multicolumn{1}{c|}{\cellcolor[HTML]{FFFFFF}\textbf{ATE (m)}} 
& \multicolumn{1}{c|}{\cellcolor[HTML]{FFFFFF}\textbf{ATE (\%)}} 
& \multicolumn{1}{c|}{\cellcolor[HTML]{FFFFFF}\textbf{ATE (m)}} 
& \multicolumn{1}{c|}{\cellcolor[HTML]{FFFFFF}\textbf{ATE (\%)}} 
& \multicolumn{1}{c|}{\cellcolor[HTML]{FFFFFF}\textbf{ATE (m)}} 
& \textbf{ATE (\%)} 
& \multicolumn{1}{c|}{\cellcolor[HTML]{FFFFFF}\textbf{ATE (m)}} 
& \multicolumn{1}{c|}{\cellcolor[HTML]{FFFFFF}\textbf{ATE (\%)}} 
& \multicolumn{1}{c|}{\cellcolor[HTML]{FFFFFF}\textbf{ATE (m)}} & \textbf{ATE (\%)} \\ \hline
\textbf{ISEC\_Lab1}    
& \multicolumn{1}{c|}{\cellcolor[HTML]{FFFFFF}2.65}             
& \multicolumn{1}{c|}{\cellcolor[HTML]{FFFFFF}1.74}              
& \multicolumn{1}{c|}{\cellcolor[HTML]{FFFFFF}1.23}             
& \multicolumn{1}{c|}{\cellcolor[HTML]{FFFFFF}0.81}              
& \multicolumn{1}{c|}{\cellcolor[HTML]{FFFFFF}0.84}             
& \multicolumn{1}{c|}{\cellcolor[HTML]{FFFFFF}0.56}              
& \multicolumn{1}{c|}{\cellcolor[HTML]{FFFFFF}\textbf{0.62}}             
& \multicolumn{1}{c|}{\cellcolor[HTML]{FFFFFF} \textbf{0.41}} 
& \multicolumn{1}{c|}{\cellcolor[HTML]{FFFFFF} \textendash}                 
& \multicolumn{1}{c|}{\cellcolor[HTML]{FFFFFF}\textendash}                  
& \multicolumn{1}{c|}{\cellcolor[HTML]{FFFFFF}14.83} &  \multicolumn{1}{c|}{\cellcolor[HTML]{FFFFFF}9.75} \\
\textbf{ISEC\_Ground1} 
& \multicolumn{1}{c|}{\cellcolor[HTML]{FFFFFF}0.99}             
& \multicolumn{1}{c|}{\cellcolor[HTML]{FFFFFF}1.07}              
& \multicolumn{1}{c|}{\cellcolor[HTML]{FFFFFF}0.350}            
& \multicolumn{1}{c|}{\cellcolor[HTML]{FFFFFF}0.38}              
& \multicolumn{1}{c|}{\cellcolor[HTML]{FFFFFF}0.321}            
& \multicolumn{1}{c|}{\cellcolor[HTML]{FFFFFF}0.34}             
& \multicolumn{1}{c|}{\cellcolor[HTML]{FFFFFF}\textbf{0.296}}  
&  \multicolumn{1}{c|}{\cellcolor[HTML]{FFFFFF}\textbf{0.32}}              
& \multicolumn{1}{c|}{\cellcolor[HTML]{FFFFFF}1.13}                 
& \multicolumn{1}{c|}{\cellcolor[HTML]{FFFFFF}1.25}  
& \multicolumn{1}{c|}{\cellcolor[HTML]{FFFFFF}1.69}  
&  \multicolumn{1}{c|}{\cellcolor[HTML]{FFFFFF}1.88}   \\
\textbf{ISEC\_Ground2} 
& \multicolumn{1}{c|}{\cellcolor[HTML]{FFFFFF}2.41}             
& \multicolumn{1}{c|}{\cellcolor[HTML]{FFFFFF}2.51}              
& \multicolumn{1}{c|}{\cellcolor[HTML]{FFFFFF}2.38}             
& \multicolumn{1}{c|}{\cellcolor[HTML]{FFFFFF}2.47}              
& \multicolumn{1}{c|}{\cellcolor[HTML]{FFFFFF}1.61}             
& \multicolumn{1}{c|}{\cellcolor[HTML]{FFFFFF}1.67}              
& \multicolumn{1}{c|}{\cellcolor[HTML]{FFFFFF}\textbf{1.2}} 
& \multicolumn{1}{c|}{\cellcolor[HTML]{FFFFFF}\textbf{1.25}}              
& \multicolumn{1}{c|}{\cellcolor[HTML]{FFFFFF}2.69}                 
& \multicolumn{1}{c|}{\cellcolor[HTML]{FFFFFF}2.98}                  
& \multicolumn{1}{c|}{\cellcolor[HTML]{FFFFFF}3.20}     
& \multicolumn{1}{c|}{\cellcolor[HTML]{FFFFFF}3.55} \\
\textbf{ISEC\_floor5}  
& \multicolumn{1}{c|}{\cellcolor[HTML]{FFFFFF}0.501}                 
& \multicolumn{1}{c|}{\cellcolor[HTML]{FFFFFF}0.27}                  
& \multicolumn{1}{c|}{\cellcolor[HTML]{FFFFFF}0.322}                 
& \multicolumn{1}{c|}{\cellcolor[HTML]{FFFFFF}0.173}                  
& \multicolumn{1}{c|}{\cellcolor[HTML]{FFFFFF}0.317}                 
& \multicolumn{1}{c|}{\cellcolor[HTML]{FFFFFF}0.17}                  
& \multicolumn{1}{c|}{\cellcolor[HTML]{FFFFFF}\textbf{0.286}}   
& \multicolumn{1}{c|}{\cellcolor[HTML]{FFFFFF} \textbf{0.153}}    
& \multicolumn{1}{c|}{\cellcolor[HTML]{FFFFFF}2.10}                   
& \multicolumn{1}{c|}{\cellcolor[HTML]{FFFFFF}1.13}                  
& \multicolumn{1}{c|}{\cellcolor[HTML]{FFFFFF}1.774}  
& \multicolumn{1}{c|}{\cellcolor[HTML]{FFFFFF}0.95}   \\
\textbf{Curry\_center} 
& \multicolumn{1}{c|}{\cellcolor[HTML]{FFFFFF}1.18}             
& \multicolumn{1}{c|}{\cellcolor[HTML]{FFFFFF}1.5}               
& \multicolumn{1}{c|}{\cellcolor[HTML]{FFFFFF}0.482}            
& \multicolumn{1}{c|}{\cellcolor[HTML]{FFFFFF}0.63}              
& \multicolumn{1}{c|}{\cellcolor[HTML]{FFFFFF}0.359}            
& \multicolumn{1}{c|}{\cellcolor[HTML]{FFFFFF}0.46}              
& \multicolumn{1}{c|}{\cellcolor[HTML]{FFFFFF}\textbf{0.292}}  
& \multicolumn{1}{c|}{\cellcolor[HTML]{FFFFFF}\textbf{0.38}}              
& \multicolumn{1}{c|}{\cellcolor[HTML]{FFFFFF}1.79}                 
& \multicolumn{1}{c|}{\cellcolor[HTML]{FFFFFF}2.27}                  
& \multicolumn{1}{c|}{\cellcolor[HTML]{FFFFFF}1.423} 
& \multicolumn{1}{c|}{\cellcolor[HTML]{FFFFFF}1.8} \\
\textbf{Falmouth}      
& \multicolumn{1}{c|}{\cellcolor[HTML]{FFFFFF}210.02}               
& \multicolumn{1}{c|}{\cellcolor[HTML]{FFFFFF}8.05}              
& \multicolumn{1}{c|}{\cellcolor[HTML]{FFFFFF}48.744}           
& \multicolumn{1}{c|}{\cellcolor[HTML]{FFFFFF}1.87}              
& \multicolumn{1}{c|}{\cellcolor[HTML]{FFFFFF}42.179}           
& \multicolumn{1}{c|}{\cellcolor[HTML]{FFFFFF}1.62}              
& \multicolumn{1}{c|}{\cellcolor[HTML]{FFFFFF} \textbf{28.481}}    
& \multicolumn{1}{c|}{\cellcolor[HTML]{FFFFFF}\textbf{1.09}}   
& \multicolumn{1}{c|}{\cellcolor[HTML]{FFFFFF} x }                 
& \multicolumn{1}{c|}{\cellcolor[HTML]{FFFFFF} x }                  
& \multicolumn{1}{c|}{\cellcolor[HTML]{FFFFFF} x } 
& \multicolumn{1}{c|}{\cellcolor[HTML]{FFFFFF} x }                
\\ \hline
\end{tabular}}}
\caption{Analysis of the accuracy of our method for different OV and N-OV configurations. The results indicate a general trend of improved performance as the number of cameras increases. N-OV setups perform poorly compared to their OV  counterparts due to scale drift. The symbol `-' indicates tracking loss, while `x' denotes unavailable data.}
\label{tab:ablation_mcslam}
\vspace{-6mm}
\end{table*}

%% file: sections/results.tex
\section{results}
Here, we discuss the experimental results for several challenging indoor and outdoor trajectories. For quantitative analysis, we use Absolute Translation Error (ATE) obtained by aligning the estimated trajectory with ground truth and computing the mean error between corresponding poses, shown in tables II-IV. When ground truth trajectory is unavailable, we use a visual target at the start and end of the trajectory to compute accumulated drift. 
\subsection{Comparing with the State of the Art (SOTA) Algorithms}
We compare the performance of our method with existing open source SLAM algorithms.\\
\textbf{Stereo SLAM evaluation}:
We evaluate ORBSLAM3{\cite{campos2021orb}} and SVO Pro{\cite{Forster2017}}, two popular visual SLAM systems in stereo mode. From \mbox{\cref{tab:slam_accuracy_stereo}}, we can observe that with a stereo setup, our method exhibits more accuracy in three out of the six datasets compared to ORBSLAM3 and SVO. While the error difference is not much for ISEC\_Lab1, ISEC\_Ground1, and ISEC\_Ground2 trajectories, it is significant for the Curry\_center trajectory, which is both longer and has much more dynamic content with people moving around. We observed that SVO is more brittle with respect to dynamic content. In the Falmouth sequence, we perform poorly compared to ORBSLAM3 and SVO in the stereo configuration. This sequence, which involves off-road outdoor images, featured dried foliage resulting in poor feature extraction and hence a loss in tracking. This impacted the performance of our method, while the other methods were able to relocalize more accurately.
We disabled loop closing in ORBSLAM3 in the comparisons to compute the accumulated drift accurately. With loop closing, ORBSLAM3's estimates may improve. However, for really long trajectories like the Falmouth sequence, we observed that even with loop closing on, ORBSLAM3 could not recover from the accumulated drift.

\begin{figure}[!b]
    \centering
     \captionsetup{font={footnotesize},labelfont={bf,sf}}
    \includegraphics[width=\columnwidth]{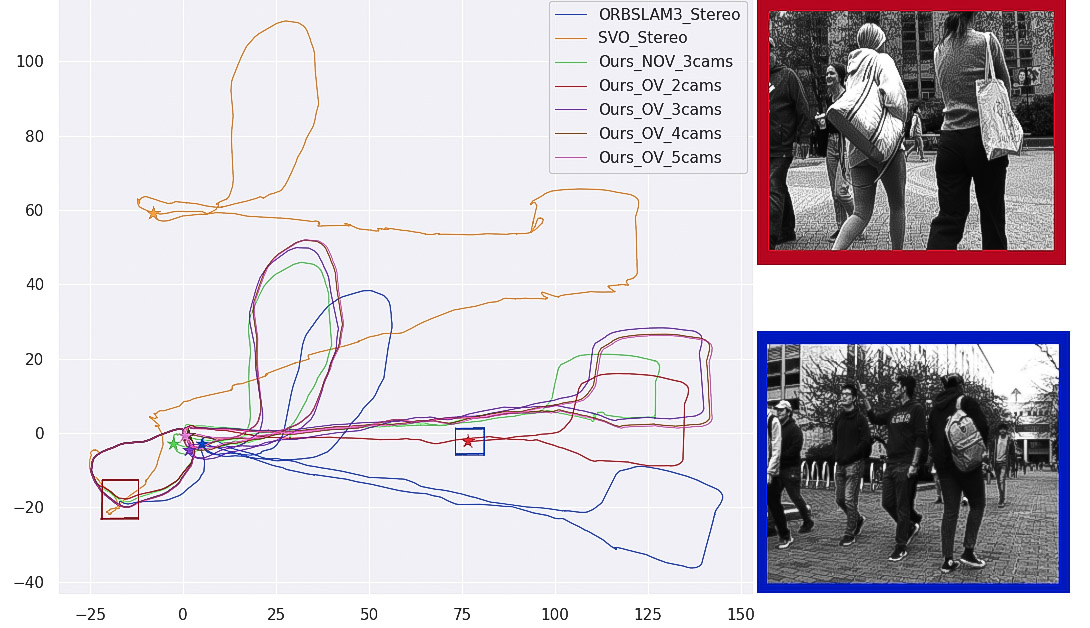}
    \caption{Estimated trajectories of the Curry\_center sequence with outdoor data and dynamic content. Stars indicate final positions of trajectory estimates. Accuracy and robustness improve with increasing number of cameras in OV configurations, as shown by accumulated drift in final position. Red and blue boxes highlight tracking failures caused by occluding dynamic objects. N-OV configuration exhibits scale issues compared to OV configuration but is robust to dynamic content.}
    \label{fig:ablation_2}
    \vspace{-2mm}
\end{figure}

\begin{figure*}[ht!]
\vspace{2mm}
\centering
\captionsetup{font={footnotesize}, labelfont={bf,sf}}
\begin{tabular}{c}
\subfloat[]{\includegraphics[clip, trim={0cm 0cm 0cm 0cm},width=0.315\linewidth, height=117pt]{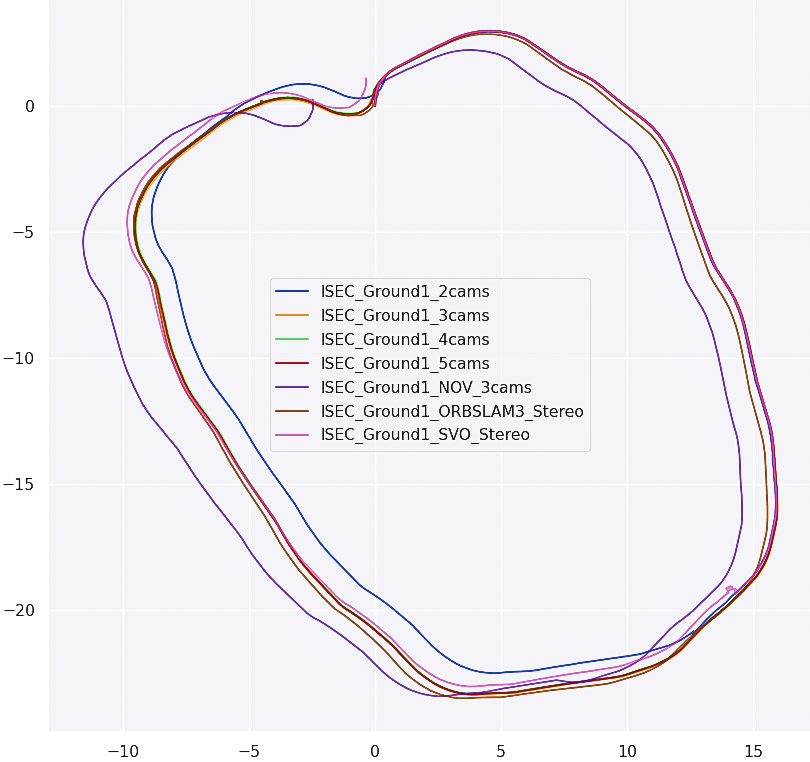}}
\subfloat[]{\includegraphics[clip, trim={0.2cm 0.45cm 0cm 0cm},width=0.61\linewidth, height=125pt]{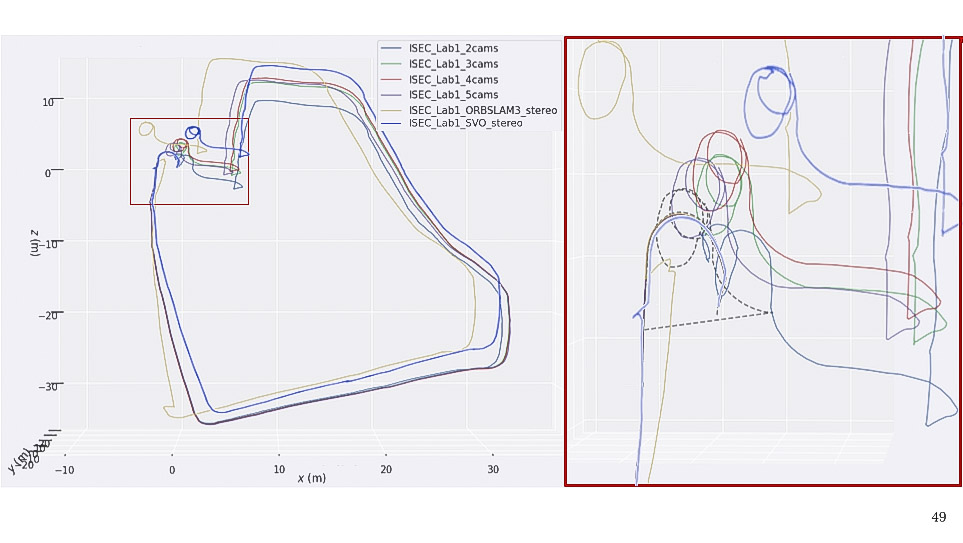}}
\end{tabular}
\caption{Estimated trajectories of the proposed SLAM system, ORBSLAM3, and SVO for ISEC\_Ground1(a) and ISEC\_Lab1 (b) sequences. In (b), the ground truth is shown as a dashed line. For ISEC Ground1 sequence(a), the robot's start and end positions are the same, facilitating performance evaluation. We achieve comparable results to ORBSLAM3 and SVO in stereo setup and demonstrate improved accuracy with increasing overlapping cameras. Notably, ORBSLAM3 and SVO encounter tracking errors due to dynamic objects along the trajectory.}
\label{fig:ablation_1}
\vspace{-4mm}
\end{figure*}

Figures {\ref{fig:ablation_1}}(a) and {\ref{fig:ablation_1}}(b) show the estimated trajectories of ISEC\_Ground1 and ISEC\_Lab1 sequences. 
In stereo mode, we perform similarly to ORBSLAM3 and SVO and outperform as we increase cameras. At several places along the trajectory, ORBSLAM3 and SVO show artifacts and tracking errors in the estimated poses due to dynamic features (also shown in {\cref{fig:ablation_2}}). We can deal with dynamic objects better as we get more support from features distributed across the FoV and are not limited by just the matched stereo features.

\begin{figure*}[ht!]
    \vspace{2mm}
    \centering
    \captionsetup{font={footnotesize}, labelfont={bf,sf}, skip=0pt}
    \hspace{-0.01\textwidth}%
    \begin{minipage}{.24\linewidth}
    \begin{subfigure}[t]{.9\linewidth}
        \includegraphics[clip, trim={2cm 1cm 1cm 1.2cm},width=\textwidth, height=100pt]{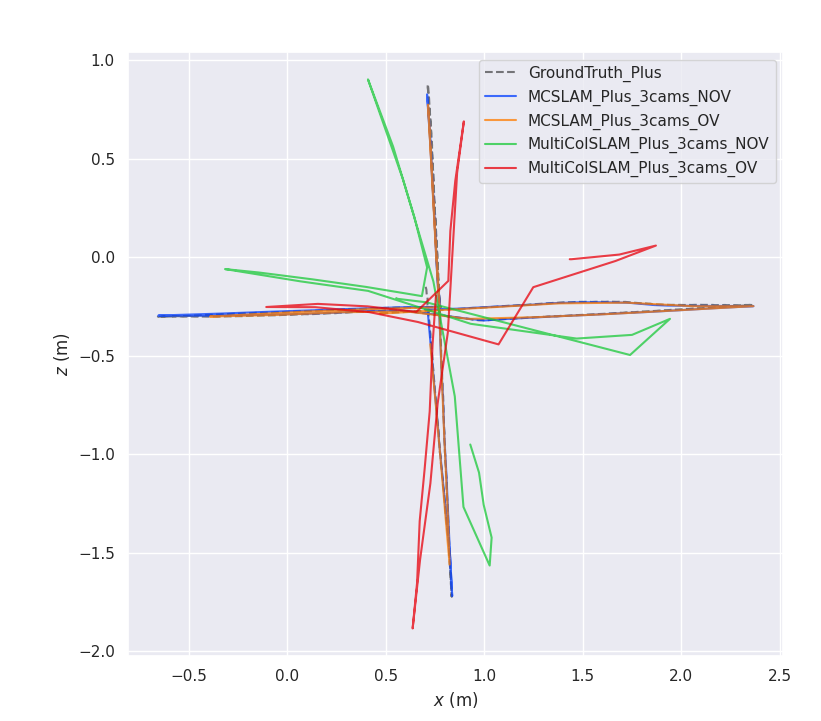}
        \caption{}
        \label{fig:plus_mcmethods}
    \end{subfigure}
    \end{minipage}
    \begin{minipage}{.24\linewidth}
        \begin{subfigure}[b]{.9\linewidth}
            \includegraphics[clip, trim={2cm 1cm 1cm 1.5cm},width=\textwidth, height=103pt]{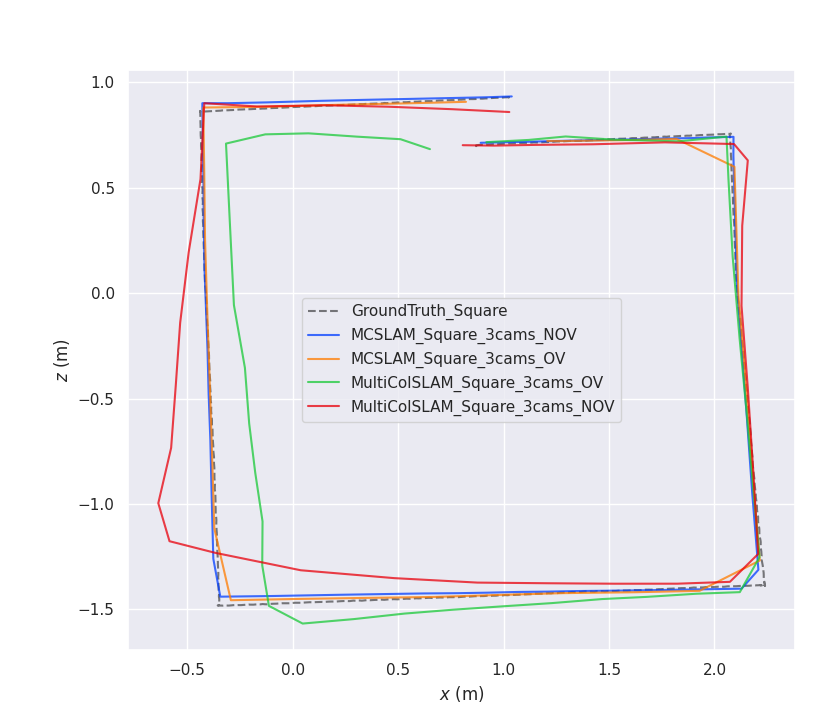}
            \caption{}
            \label{fig:square_mcmethods}
        \end{subfigure}
    \end{minipage}
    \begin{minipage}{.24\linewidth}
        \begin{subfigure}[b]{.9\linewidth}
            \includegraphics[clip,trim={2.5cm 1cm 1.5cm 1cm}, width=120pt, height=100pt] 
                {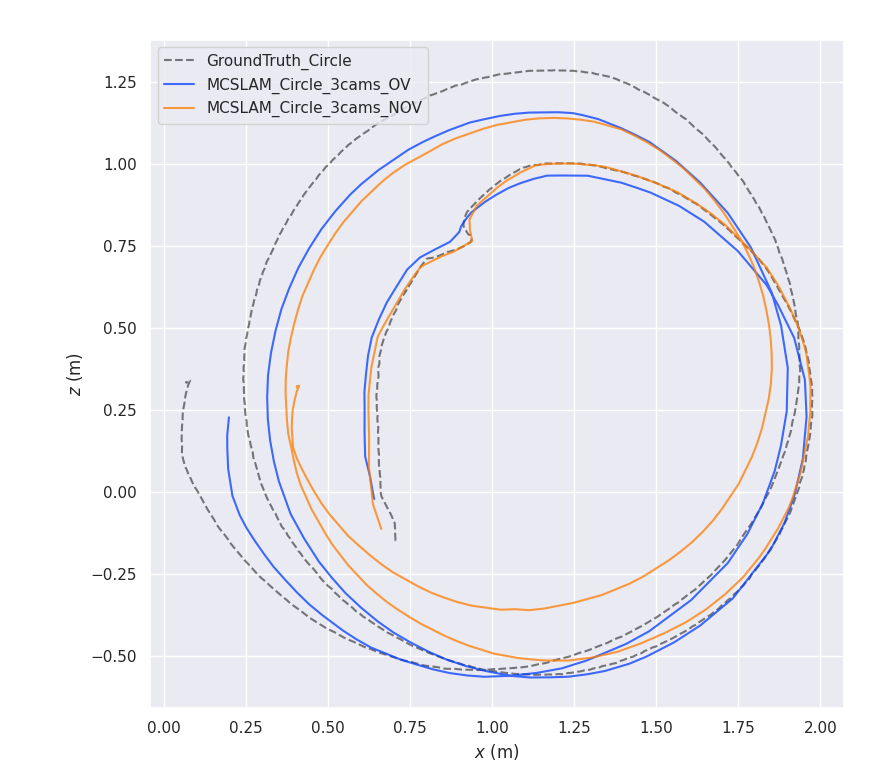}
            \caption{}
            \label{fig:intra_vs_mono}
        \end{subfigure}
    \end{minipage}
    \begin{minipage}{.24\linewidth}
        \begin{subfigure}[b]{.99\linewidth}
            \includegraphics[clip, width=\textwidth, height=100pt]{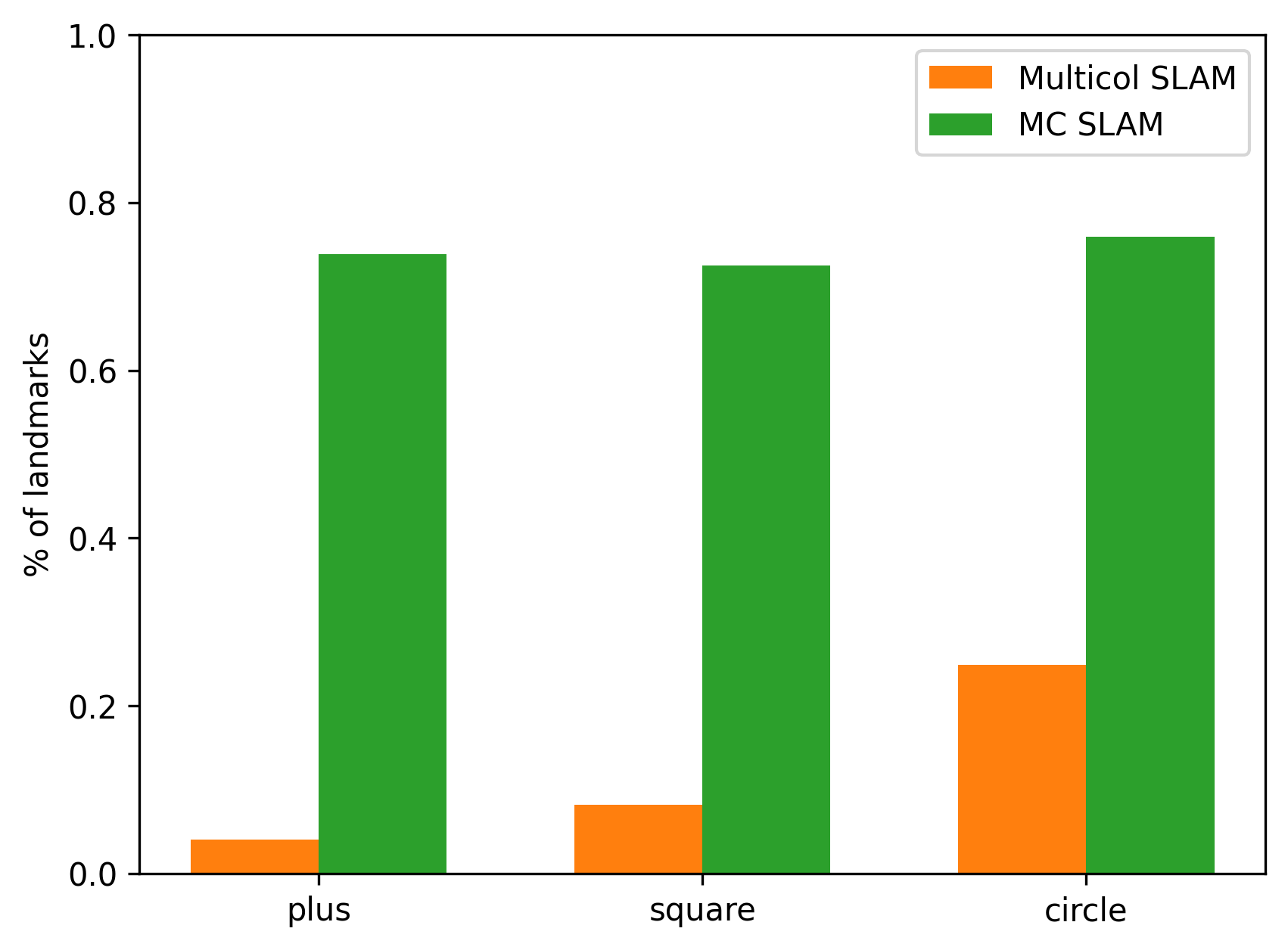}
            \caption{}
            \label{fig:intra_vs_mono}
        \end{subfigure}
    \end{minipage}

    \caption{ Performance comparisons between MultiCol SLAM and our system: (a) - (c) show estimated trajectories using a 3-camera setup with OV and N-OV configurations on the simple sequences. (d) Percentage of average no. of landmarks found in more than one component camera by both systems. Our system utilizes metric features more effectively, avoiding scale loss and maximizing feature utilization.}
    \label{fig:ablation_3}
    \vspace{-4mm}
\end{figure*}

\textbf{Multi-Camera Baseline:}
We use MultiColSLAM{\cite{urban2016multicol}} as the baseline multi-camera SLAM method and evaluate it across various OV and N-OV configurations. During our experiments, MultiColSLAM could not track well and failed on all of our datasets shown in \mbox{\cref{tab:dataset_desc}}, irrespective of the multi-camera setup used. For a fair comparison, we collected an additional set of sequences by tracing simple shapes such as pluses, squares, and circles in a feature-rich static environment with Optitrack ground truth. 
The evaluation results in \mbox{\cref{tab:mcslam_compare}} show that our method outperforms the baseline in all the sequences. We observed that MultiColSLAM is sensitive to rotational motion and experiences continuous tracking failures for the circle sequence. It works on plus and square sequences but displays significant scale errors even when the cameras overlap. This is because the features are matched, tracked, and mapped independently in each camera without leveraging the camera arrangement to get metric features for scale. Figure {\ref{fig:ablation_3}} shows estimated trajectories for 3-camera OV and N-OV setups and the percentage of features that are observed in more than one component camera under the completely overlapping case.

\subsection{Effect of Camera Configuration}
Along with SOTA comparisons, we also evaluate and discuss the effect of the following parameters on the performance of the proposed SLAM pipeline.
    
\subsubsection{Accuracy}

\textbf{Number of Cameras:} Within the overlapping configuration, we evaluate our method by choosing a subset of cameras and increasing the number of cameras for each trial. We start with 2 cameras (stereo) with the smallest baseline and go up to 5 cameras in the front-facing array. Table\ref{tab:ablation_mcslam} shows that within each sequence, the ATE decreases as the number of overlapping cameras increases. We can see the same trend in the trajectory plots shown in \cref{fig:ablation_1}(a), \cref{fig:ablation_1}(b) and \cref{fig:ablation_2}. The estimated trajectory is closer to ground truth as we increase the cameras as observed in the zoomed part of \cref{fig:ablation_1}(b).\\
\setlength{\belowcaptionskip}{-1pt}

\textbf{Overlap vs Non-Overlap:}
From table\ref{tab:ablation_mcslam}, errors for the N-OV configuration are greater than the OV configurations in most cases. For instance, the accuracy of the set of 3 overlapping cameras chosen from the front-facing array is higher than the set of the same number of non-overlapping cameras facing different directions.  This is because the N-OV setup quickly accumulates scale drift. The error is particularly high in the ISEC\_Lab1 sequence, which has narrow featureless corridors and reflective glass walls, rendering the sideward-looking cameras useless for tracking. 

\subsubsection{Robustness} To study the robustness of tracking across different camera configurations, we take a closer look at multiple runs of the SLAM pipeline for Curry\_center sequence, which is a large dataset (597m) that features heavy dynamic content as shown in \cref{fig:ablation_2}. This data was collected by driving the robot around campus on a regular working day while there was a lot of human activity. Figure\ref{fig:ablation_2} shows the best run for each camera configuration. The red and blue boxes indicate the places where we experience tracking failures, and the corresponding images acquired by one of the cameras are displayed. In our runs, we experience the highest number of tracking failures in the two-camera (stereo) OV configuration. The tracking failures also happen in the 3-camera OV configuration but are less frequent than in the two-camera case. The OV configuration with 4 and 5 cameras runs successfully, closely following each other. The N-OV 3-camera setup does not fail in the presence of dynamic objects because when one view is occluded, it has the support of the other views to track features. However, the trajectory estimates are inaccurate and exhibit severe scale errors.

\captionsetup[subfigure]{skip=0pt}

\subsection{Run Time Performance}
We conclude the evaluation by measuring the average time taken to process a single multi-camera frame across various camera configurations. Table\ref{tab:slam_runtime} reports individual processing times for feature extraction, tracking and mapping, back-end optimization modules, and the total processing time per frame. The results show that the processing time increases with the number of cameras in the OV configuration. This is expected because we have the extra burden of computing multi-view features across cameras in the front end. The computational load in the back end also increases due to increased observations. We can achieve a maximum processing speed of 19.1 fps for a stereo configuration and a minimum of 11.45 fps for five cameras in the OV configuration.
\input{tables/run_time}

%% file: tables/run_time.tex
\begin{table}
\vspace{2mm}
\centering
\captionsetup{font=footnotesize,labelfont={bf,sf}}
\resizebox{\columnwidth}{!}{%
\begin{tabular}{|llllll|}
\hline
\multicolumn{6}{|c|}{\textbf{Average Time Taken for Processing (Milli Seconds)}} \\ \hline
\multicolumn{1}{|l|}{} &
  \multicolumn{1}{l|}{\textbf{2 cams}} &
  \multicolumn{1}{l|}{\textbf{3cams}} &
  \multicolumn{1}{l|}{\textbf{4cams}} &
  \multicolumn{1}{l|}{\textbf{5cams}} &
  \textbf{N-OV} \\ \hline
\multicolumn{1}{|l|}{\textbf{Feature Extraction}} &
  \multicolumn{1}{l|}{23.1} &
  \multicolumn{1}{l|}{25.61} &
  \multicolumn{1}{l|}{29.16} &
  \multicolumn{1}{l|}{32.64} &
  21.3 \\ \hline
\multicolumn{1}{|l|}{\textbf{Tracking \& Mapping}} &
  \multicolumn{1}{l|}{8.02} &
  \multicolumn{1}{l|}{8.64} &
  \multicolumn{1}{l|}{9.85} &
  \multicolumn{1}{l|}{10.78} &
  8.32 \\ \hline
\multicolumn{1}{|l|}{\textbf{Optimization}} &
  \multicolumn{1}{l|}{10.23} &
  \multicolumn{1}{l|}{20.7} &
  \multicolumn{1}{l|}{27.02} &
  \multicolumn{1}{l|}{31.83} &
  10.99 \\ \hline
\multicolumn{1}{|l|}{\textbf{Total Avg time}} &
  \multicolumn{1}{l|}{52.14} &
  \multicolumn{1}{l|}{64.91} &
  \multicolumn{1}{l|}{74.89} &
  \multicolumn{1}{l|}{87.33} &
  58.08 \\ \hline
\end{tabular}%
}
\caption{Run time performance of different steps in the SLAM pipeline. The processing time increases with number of cameras. We achieve 11 fps for 5-camera OV configuration which is close to real-time.}
\label{tab:slam_runtime}
\vspace{-1mm}
\end{table}

%% file: sections/conclusion.tex
\section{Conclusion}
We presented a generic multi-camera SLAM framework that can adapt to any arbitrary camera configuration. The core contribution is the camera configuration independent design and real-time implementation across the complete SLAM pipeline. Leveraging the camera geometry, we extract well-distributed multi-view features by effectively using the overlapping FoVs among cameras. We conducted extensive evaluations on real-world datasets collected using a custom-built camera rig featuring a variety of challenging conditions. We also benchmark the performance of the SLAM pipeline in terms of the number of cameras and the overlap information that define the camera configuration. This analysis can be used in designing multi-camera systems for accurate and robust SLAM. This work addresses the gap between the state-of-the-art visual SLAM algorithms and their applicability to real-world deployments of multi-camera systems. We make the code and datasets publicly available to foster research in this direction.

